%% file: main.tex
\documentclass[10pt,twocolumn,letterpaper]{article}
\usepackage{cvpr}

\usepackage{dsfont}
\usepackage{times}
\usepackage{amsmath,amssymb} 
\usepackage{graphicx}
\usepackage{caption}
\usepackage{subcaption}

\usepackage[pagebackref=true,breaklinks=true,letterpaper=true,colorlinks,bookmarks=false]{hyperref}

\cvprfinalcopy 

\def\cvprPaperID{1191} 

\ifcvprfinal\fi
\begin{document}

\title{Unsupervised Event-based Learning of Optical Flow, Depth, and Egomotion}

%


\author{Alex Zihao Zhu, 
Liangzhe Yuan, 
Kenneth Chaney, 
Kostas Daniilidis\\
University of Pennsylvania\\
\{alexzhu, lzyuan, chaneyk, kostas\}@seas.upenn.edu
}

\maketitle
\global\csname @topnum\endcsname 0
\global\csname @botnum\endcsname 0
\begin{abstract}
In this work, we propose a novel framework for unsupervised learning for event cameras that learns motion information from only the event stream. In particular, we propose an input representation of the events in the form of a discretized volume that maintains the temporal distribution of the events, which we pass through a neural network to predict the motion of the events. This motion is used to attempt to remove any motion blur in the event image. We then propose a loss function applied to the motion compensated event image that measures the motion blur in this image. We train two networks with this framework, one to predict optical flow, and one to predict egomotion and depths, and evaluate these networks on the Multi Vehicle Stereo Event Camera dataset, along with qualitative results from a variety of different scenes.\end{abstract}

\input{tex/TitleFig.tex}
\input{tex/ArchitectureFigure.tex}
\input{tex/Introduction.tex}	
\input{tex/Qualitative.tex}
\input{tex/RelatedWork.tex}
\input{tex/Method.tex}
\input{tex/ScaledDepthTable.tex}
\input{tex/Experiments.tex}
\input{tex/Results.tex}
\input{tex/FlowFigs.tex}
\section{Conclusions}
In this work, we demonstrate a novel input representation for event cameras, which, when combined with our motion compensation based loss function, allows a deep neural network to learn to predict optical flow and ego-motion and depth from the event stream only.
\section{Acknowledgements}
Thanks to Tobi Delbruck and the team at iniLabs and iniVation for providing and supporting the DAVIS-346b cameras. This work was supported in part by the Semiconductor Research Corporation (SRC)
and DARPA. We also gratefully appreciate support through the following grants: NSF-DGE-0966142 (IGERT), NSF-IIP-1439681 (I/UCRC), NSF-IIS-1426840, NSF-IIS-1703319, NSF MRI 1626008, ARL RCTA W911NF-10-2-0016, ONR N00014-17-1-2093, the Honda Research Institute and the DARPA FLA program.

{\small
\bibliographystyle{ieee}
\bibliography{refs}
}
\end{document}

%% file: tex/TitleFig.tex
\begin{figure}
    \centering
    \includegraphics[width=0.45\textwidth]{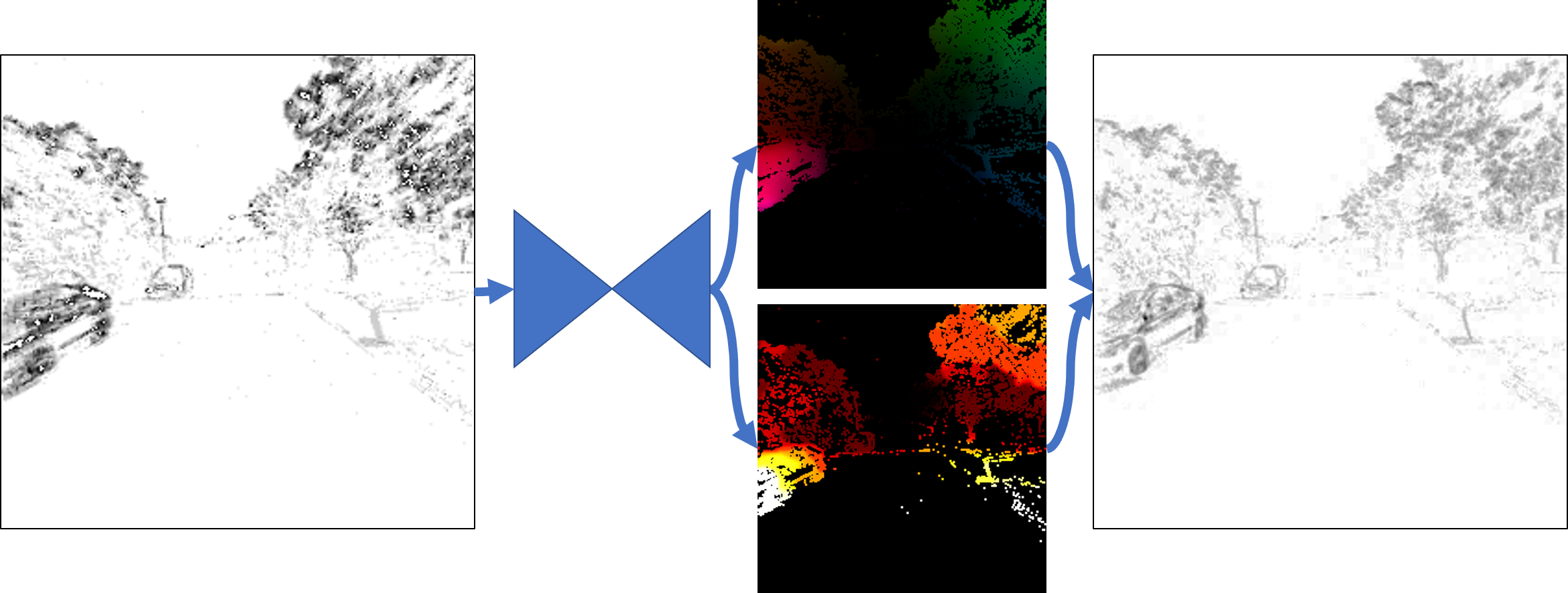}
    \caption{Our network learns to predict motion from motion blur by predicting optical flow (top) or egomotion and depth (bottom) from a set of input, blurry, events from an event camera (left), and minimizing the amount of motion blur after deblurring with the predicted motion to produce the deblurred image (right). Best viewed in color.}
    \label{fig:header}
\end{figure}

%% file: tex/ArchitectureFigure.tex
\begin{figure*}[t!]
    \centering
    \includegraphics[width=0.7\textwidth]{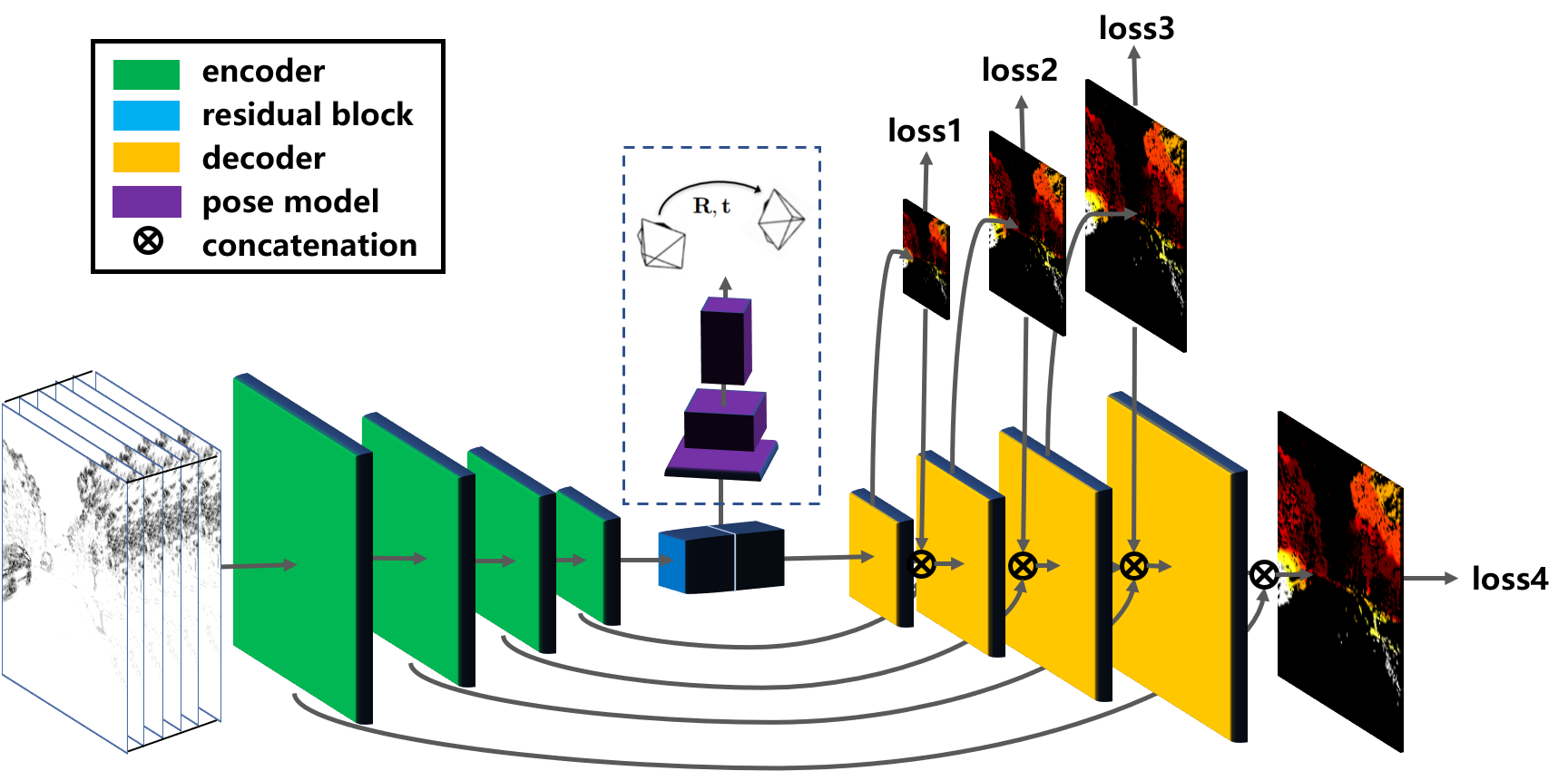}
    \caption{Network architecture for both the optical flow and egomotion and depth networks. In the optical flow network, only the encoder-decoder section is used, while in the egomotion and depth network, the encoder-decoder is used to predict depth, while the pose model predicts the egomotion. At training time, the loss is applied at each stage of the decoder, before being concatenated into the next stage of the network.}
    \label{fig:architecture}
\end{figure*}

%% file: tex/Introduction.tex
\section{Introduction}
\label{sec:intro}
Event cameras are a neuromorphically inspired, asynchronous sensing modality, that detect changes in log light intensity. When a change is detected in a pixel, the camera immediately returns an event, $e=\{x,y,t,p\}$, consisting of the position of the pixel, $x, y$, timestamp of the change, $t$, accurate to microseconds, and the polarity of the change, $p$, corresponding to whether the pixel became brighter or darker. The asynchronous nature of the camera, and the tracking in the log image space, provide numerous benefits over traditional frame based cameras, such as extremely low latency for tracking very fast motions, very high dynamic range, and significantly lower power consumption.

However, the novel output of the cameras provide new challenges in algorithm development. As the events simply reflect whether a change has occurred at a given pixel, a model of photoconsistency, as used traditional motion estimation tasks such as optical flow or structure from motion (SFM), applied directly on the events is no longer valid. As a result, there has been a significant research drive to develop new algorithms for event cameras to solve these traditional robotics problems.

There have been recent works by Zhu et al.~\cite{zhu2018ev} and Ye et al.~\cite{ye2018unsupervised} that train neural networks to learn to estimate these motion tasks in a self and unsupervised manner. These networks abstract away the difficult problem of modeling and algorithm development. However, both works still rely on photoconsistency based principles, applied to the grayscale image and an event image respectively, and, as a result, the former work relies on the presence of grayscale images, while the latter's photoconsistency assumption may not hold valid in very blurry scenes. In addition, both works take inputs that attempt to summarize the event data, and as a result lose temporal information.

In this work, we resolve these deficiencies by proposing a novel input representation that captures the full spatiotemporal distribution of the events, and a novel set of unsupervised loss functions that allows for efficient learning of motion information from only the event stream. Our input representation, a discretized event volume, discretizes the time domain, and then accumulates events in a linearly weighted fashion similar to interpolation. This representation encodes the distribution of all of the events within the spatiotemporal domain. We train two networks to predict optical flow and ego-motion and depth, and use the predictions to attempt to remove the motion blur from the event stream, as visualized in Fig.~\ref{fig:header}. Our unsupervised loss then measures the amount of motion blur in the corrected event image, which provides a training signal to the network. In addition, our deblurred event images are comparable to edge maps, and so we apply a photometric stereo loss on the census transform of these images to allow our network to learn metric poses and depths.

We evaluate both methods on the Multi Vehicle Stereo Event Camera dataset~\cite{zhu2018multi}\cite{zhu2018ev}, and compare against the equivalent grayscale based methods, as well as the prior state of the art by \cite{zhu2018ev}.

Our contributions can be summarized as:
\begin{itemize}
\item
A novel discretized event volume representation for passing events into a neural network.
\item
A novel application of a motion blur based loss function that allows for unsupervised learning of motion information from events only.
\item
A novel stereo photometric loss applied on the census transform of a pair of deblurred event images.
\item
Quantitative evaluations on the Multi Vehicle Stereo Event Camera dataset \cite{zhu2018multi}, with qualitative and quantitative evaluations from a variety of night time and other challenging scenes.
\end{itemize}

%% file: tex/Qualitative.tex
\begin{figure*}[t]
    \centering
      \includegraphics[width=0.2\textwidth]{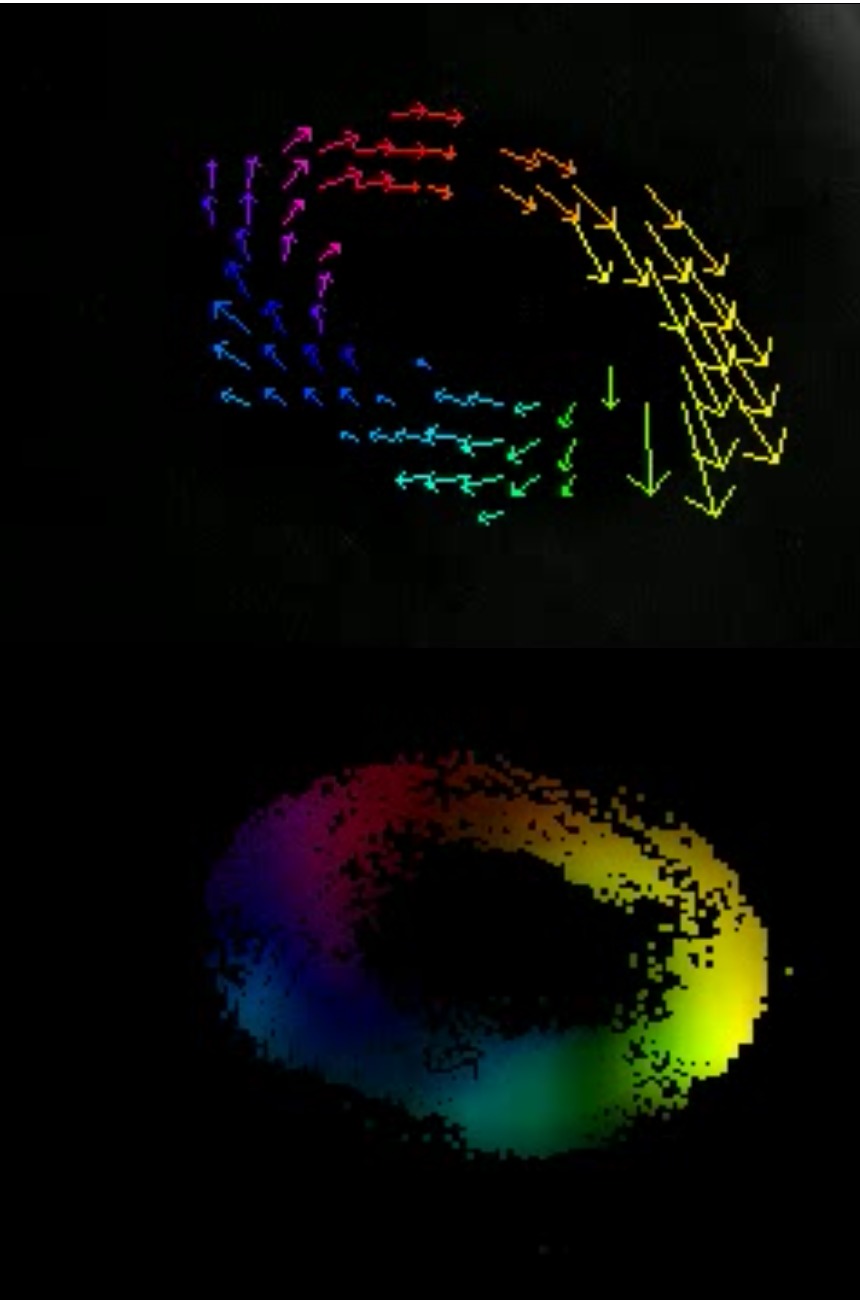}
     \includegraphics[width=0.2\textwidth]{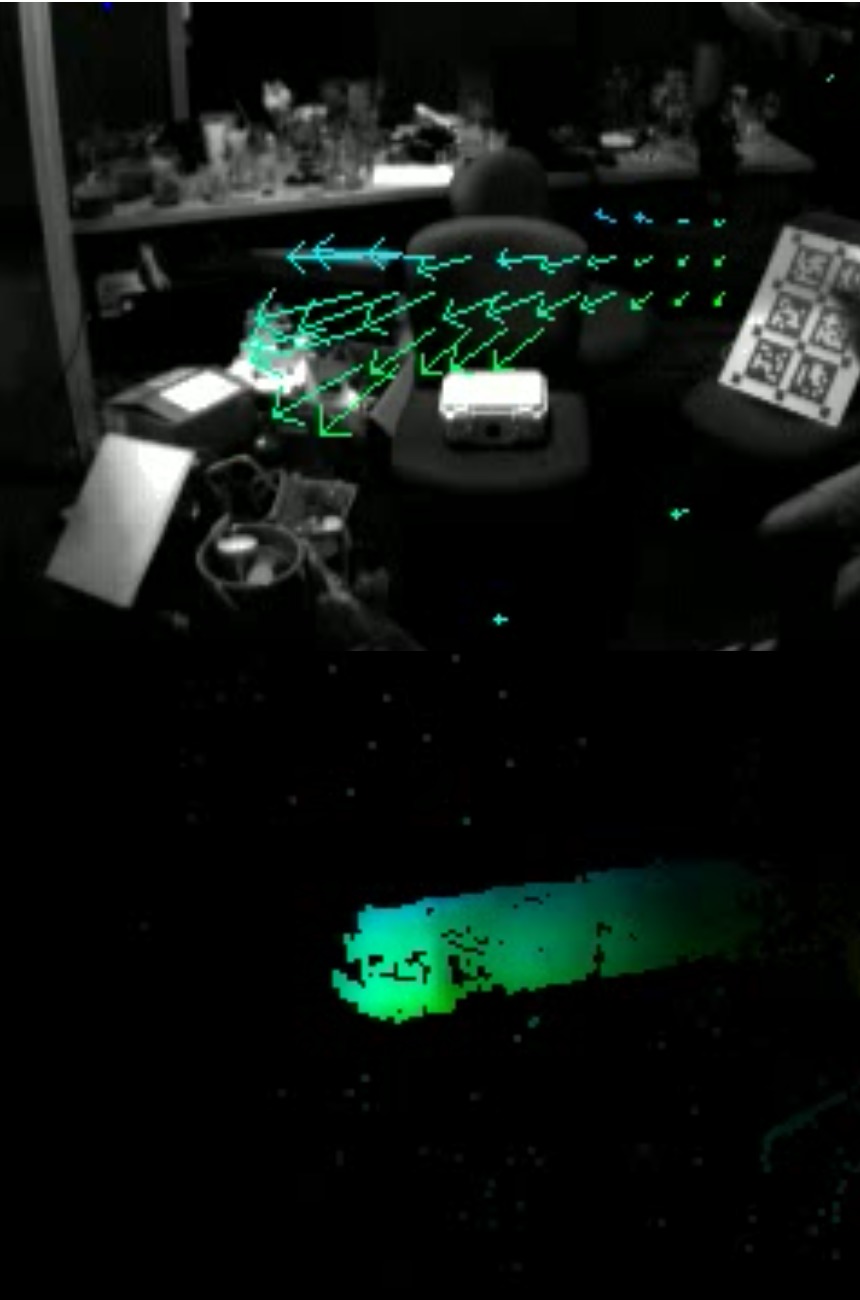}
     \includegraphics[width=0.2\textwidth]{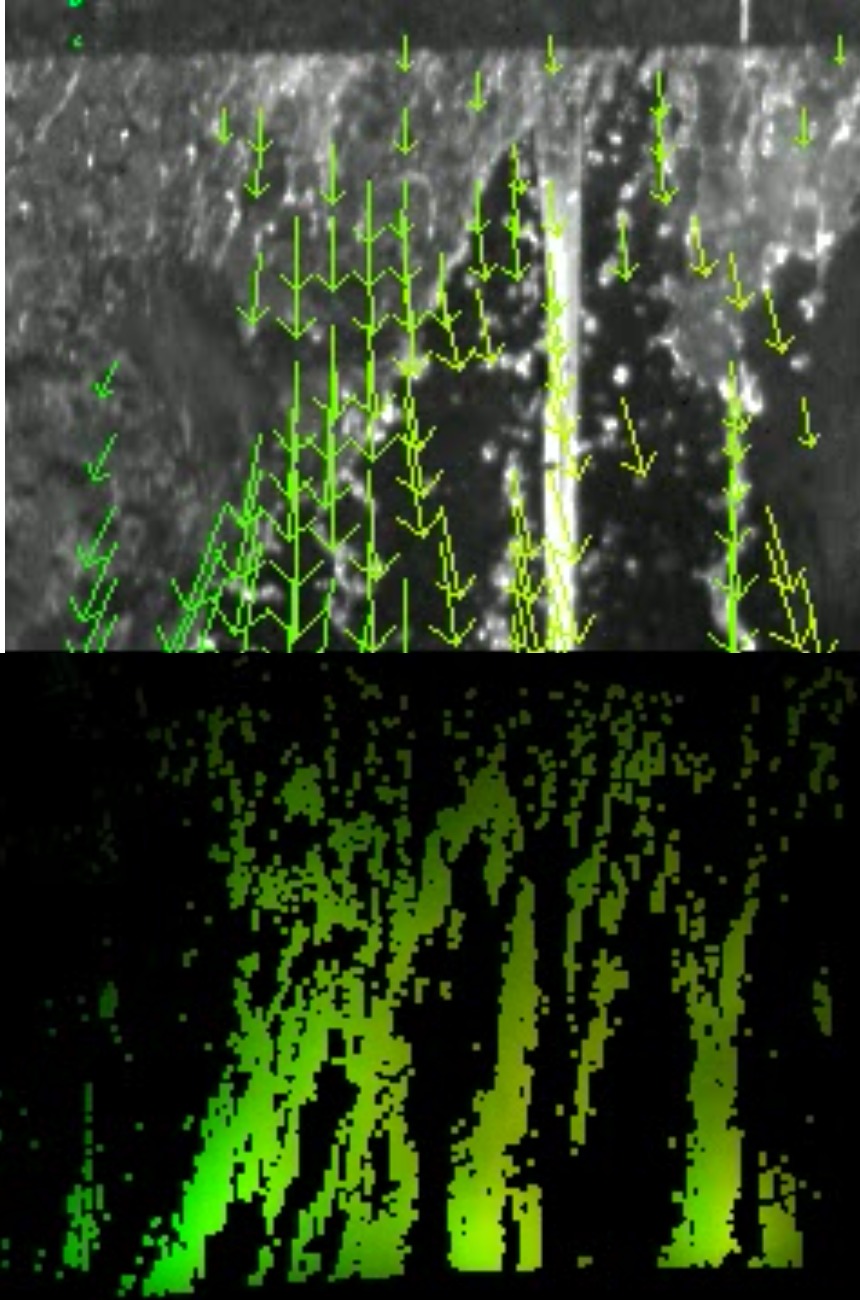}
     \caption{Our flow network is able to generalize to a variety of challenging scenes. Top images are a subset of flow vectors plotted on top of the grayscale image from the DAVIS camera, bottom images are the dense flow output of the network at pixels with events, colored by the direction of the flow. Left to right: Fidget spinner spinning at 13 rad/s in a very dark environment. Ball thrown quickly in front of the camera (the grayscale image does not pick up the ball at all). Water flowing outdoors.}
     \label{fig:generalization}
\end{figure*}

%% file: tex/RelatedWork.tex
\section{Related Work}
\label{sec:related_work}
Since the introduction of event cameras, such as Lichtsteiner et al.~\cite{lichtsteiner2008128}, there has been a strong interest in the development of algorithms that leverage the benefits provided by these cameras. In the work of optical flow, Benosman et al.~\cite{benosman2014event} showed that normal flow can be estimated by fitting a plane to the events in x-y-t space. Bardow et al.~\cite{bardow2016simultaneous} show that flow estimation can be written as a convex optimization problem that solves for the image intensity and flow jointly. 

In the space of SFM and visual odometry, Kim et al.~\cite{kim2016real} demonstrate that a Kalman filter can reconstruct the pose of the camera and a local map. Rebecq et al.~\cite{rebecq2017real} similarly build a 3D map, which they localize from using the events. Zhu et al.~\cite{zhu2017evio} use an EM based feature tracking method to perform visual-inertial odometry, while Rebecq et al.~\cite{rebecq2017real} use motion compensation to deblur the event image, and run standard image based feature tracking to perform visual-inertial odometry. 

For model-free methods, self-supervised and unsupervised learning have allowed deep networks to learn motion and the structure of a scene, using only well established geometric principles. Yu et al.~\cite{jason2016back} established that a network can learn optical flow from brightness constancy with a smoothness prior, while Maqueda et al.~\cite{maqueda2018event} extend this work by applying a bidirectional census loss to improve the quality of the flow. In a similar fashion, Zhou et al.~\cite{zhou2017unsupervised} show that a network can learn a camera's egomotion and depth using camera reprojection and a photoconsistency loss. Zhan et al.~\cite{zhan2018unsupervised} and Vijayanarasimhan et al.~\cite{vijayanarasimhan2017sfm} add in a stereo constraint, allowing the network to learn absolute scale, while Wang et al.~\cite{wang2018recurrent} apply this concept with a recurrent neural network.

Recently, there have been several works, such as \cite{gallego2018unifying, gallego2017accurate, mitrokhin2018event, zhu2018ev, zhu2018realtime}, that have shown that optical flow, and other types of motion information, can be estimated from a spatiotemporal volume of events, by propagating the events along the optical flow direction, and attempting to minimize the motion blur in the event image. This concept of motion blur as a loss can be seen as an analogy to the photometric error in frames, as applied to events. In this work, we adapt a novel formulation of this loss from Mitrokhin et al.~\cite{mitrokhin2018event} for a neural network, by generating a single fully differentiable loss function that allows our networks to learn optical flow and structure from motion in an unsupervised manner.

%% file: tex/Method.tex
\section{Method}
\label{sec:method}
Our pipeline consists of a novel volumetric representation of the events, which we describe in Sec.~\ref{sec:input}, which is passed through a fully convolutional neural network to predict flow and/or egomotion and depth. We then use the predicted motion to try to deblur the events, and apply a loss that minimizes the amount of blur in the deblurred image, as described in Sec.~\ref{sec:loss}. This loss can be directly applied to our optical flow network, Sec.~\ref{sec:flow_network}. For the egomotion and depth network, we describe the conversion to optical flow in Sec.~\ref{sec:egomotion_loss}, as well as a novel stereo disparity loss in Sec.~\ref{sec:disp_loss}. Our architecture is summarized in Fig.~\ref{fig:architecture}.

\subsection{Input: The Discretized Event Volume}
\label{sec:input}
Selecting the appropriate input representation of a set of events for a neural network is still a challenging problem. Prior works such as Moeys et al.~\cite{moeys2016steering} and Maqueda et al.~\cite{maqueda2018event} generate an event image by summing the number of events at each pixel. However, this discards the rich temporal information in the events, and is susceptible to motion blur. Zhu et al.~\cite{zhu2018ev} and Ye et al.~\cite{ye2018unsupervised} propose image representations of the events, that summarize the number of events at each pixel, as well as the last timestamp and average timestamp at each pixel, respectively. Both works show that this is sufficient for a network to predict accurate optical flow. While this maintains some of the temporal information, a lot of information is still lost by summarizing the high resolution temporal information in the events.

In this work, we propose a novel input representation generated by discretizing the time domain. In order to improve the resolution along the temporal domain beyond the number of bins, we insert events into this volume using a linearly weighted accumulation similar to bilinear interpolation.

Given a set of $N$ input events $\{(x_i, y_i, t_i, p_i)\}_{i\in[1,N]}$, and a set $B$ bins to discretize the time dimension, we scale the timestamps to the range $[0, B-1]$, and generate the event volume as follows:
\begin{align}
t^*_i =& (B-1)(t_i - t_0) / (t_{N} - t_1)\\
V(x,y,t)=&\sum_{i} p_i k_b(x-x_i)k_b(y-y_i)k_b(t-t^*_i)\\
k_b(a) =& \max(0, 1-|a|)
\end{align}
where $k_b(a)$ is equivalent to the bilinear sampling kernel defined in Jaderberg et al. \cite{jaderberg2015spatial}. Note that the interpolation in the $x$ and $y$ dimensions is necessary when camera undistortion or rectification is performed, resulting in non integer pixel positions.

In the case where no events overlap between pixels, this representation allows us to reconstruct the exact set of events. When multiple events overlap on a voxel, the summation does cause some information to be lost, but the resulting volume still retains the distribution of the events across both the spatial and temporal dimensions within the window.

In this work, we treat the time domain as channels in a traditional 2D image, and perform 2D convolution across the $x,y$ spatial dimensions. We have tested a network with full 3D convolutions across this volume, but found negligible performance increases for a significant increase in processing time.
\input{tex/DeblurPic.tex}
\subsection{Supervision through Motion Compensation}
\label{sec:loss}
As event cameras register changes in log intensity, the standard model of photoconsistency does not directly apply onto the events. Instead, several works have applied the concept of motion compensation, as described in Rebecq et al.~\cite{rebecq2017real}, as a proxy for photoconsistency when estimating motion from a set of events. The goal of motion compensation is to use the motion model of each event to deblur the event image, as visualized in Fig.~\ref{fig:deblurring}.

For the most general case of per pixel optical flow, $u(x,y), v(x,y)$, we can propagate the events, $\{(x_i,y_i,t_i,p_i)\}_{i=1,\dots,N}$, to a single time $t'$:
\begin{align}
\begin{pmatrix}x_i'\\y_i'\end{pmatrix}=&\begin{pmatrix}x_i\\y_i\end{pmatrix}+(t'-t_i)\begin{pmatrix}u(x_i,y_i)\\v(x_i,y_i)\end{pmatrix}\label{eq:event_prop}
\end{align}
If the input flow is correct, this has the effect of reversing the motion in the events, and removing the motion blur, while for an incorrect flow, this will likely induce further motion blur.

We use a measure of the quality of this deblurring effect as the main supervision for our network. Gallego et al.~\cite{gallego2018unifying} proposed using the image variance on an image generated by the propagated events. However, we found that the network would easily overfit to this loss, by predicting flow values that push all events within each region of the image to a line. This effect is discussed further in the supplemental. Instead, we adopt the loss function described by Mitrokhin et al.~\cite{mitrokhin2018event}, who use a loss function that minimizes the sum of squares of the average timestamp at each pixel.

However, the previously proposed loss function is non-differentiable, as the authors rounded the timestamps to generate an image. To resolve this, we generate the average timestamp image using bilinear interpolation. We apply the loss by first separating the events by polarity and generating an image of the average timestamp at each pixel for each polarity, $T_{+}, T_{-}$:
\begin{align}
T_{p'}(x, y|t') =& \frac{\sum_i \mathds{1}(p_i=p') k_b(x-x_i')k_b(y-y_i')t_i}{\sum_i \mathds{1}(p_i=p') k_b(x-x_i')k_b(y-y_i')}\label{eq:timestamp_image}\\
p'\in& \{+, -\}\nonumber\\
k_b(a)=&\max(0, 1-|a|)
\end{align}
The loss is, then, the sum of the two images squared.
\begin{align}
\mathcal{L}_{\text{time}}(t')=&\sum_x \sum_y T_{+}(x, y|t')^2 + T_{-}(x, y|t')^2
\end{align}
However, using a single $t'$ for this loss poses a scaling problem. In \eqref{eq:event_prop}, the output flows, $u, v$, are scaled by $(t'-t_i)$. During backpropagation, this will weight the gradient over events with timestamps further from $t'$ higher, while events with timestamps very close to $t'$ are essentially ignored. To mitigate this scaling, we compute the loss both backwards and forwards, with $t'=0$ and $t'=t_{N-1}$, allowing all of the events to contribute evenly to the loss:
\begin{align}
\mathcal{L}_{\text{time}} =& \mathcal{L}_{\text{time}}(t_0) + \mathcal{L}_{\text{time}}(t_{N-1})\label{eq:temporal_loss}
\end{align}
Note that changing the target time, $t'$, does not change the timestamps used in \eqref{eq:timestamp_image}.

This loss function is similar to that of Benosman et al.~\cite{benosman2014event}, who model the events with a function $\Sigma_{e_i}$, such that $\Sigma_{e_i}(\mathbf{x}_i)=t_i$. In their work, they assume that the function is locally linear, and solve the minimization problem by fitting a plane to a small spatiotemporal window of events. Indeed, we can see that the gradient of the average timestamp image, $(dt/dx, dt/dy)$, corresponds to the inverse of the flow, if we assume that all events at each pixel have the same flow.

\input{tex/FlowResultsTable.tex}

\subsection{Optical Flow Prediction Network}
\label{sec:flow_network}
Using the input representation and loss described in Sec.~\ref{sec:input} and \ref{sec:loss}, we train a neural network to predict optical flow. We use an encoder-decoder style network, as in \cite{zhu2018ev}. The network outputs flow values in units of pixels/bin, which we apply to \eqref{eq:event_prop}, and eventually compute \eqref{eq:full_loss}.

Our flow network uses the temporal loss in \eqref{eq:temporal_loss}, combined with a local smoothness regularization:
\begin{align}
\mathcal{L}_{\text{smooth}}=\sum_{\vec{x}} \sum_{\vec{y}\in \mathcal{N}(\vec{x})} &\rho(u(\vec{x}) - u(\vec{y})) + \rho(v(\vec{x}) - v(\vec{y})) \label{eq:smoothness}\\
\rho(x)=&\sqrt{x^2+\epsilon^2}\label{eq:charbonnier}
\end{align}
where $\rho(x)$ is the Charbonnier loss function~\cite{charbonnier1994two}, and $\mathcal{N}(x, y)$ is the 4-connected neighborhood around $(x, y)$.

The total loss for the flow network is:
\begin{align}
\mathcal{L}_{\text{flow}}=&\mathcal{L}_{\text{time}} + \lambda_1 \mathcal{L}_{\text{smooth}} \label{eq:full_loss}
\end{align}

\subsection{Egomotion and Depth Prediction Network}
\label{sec:sfm_network}
We train a second network to predict the egomotion of the camera and the structure of the scene, in a similar manner to \cite{zhan2018unsupervised, vijayanarasimhan2017sfm}. Given a pair of time synchronized discretized event volumes from a stereo pair, we pass each volume into our network separately, but use both at training time to apply a stereo disparity loss, allowing our network to learn metric scale. We apply a temporal timestamp loss defined in Sec. \ref{sec:loss}, and a robust similarity loss between the census transforms \cite{zabih1994non, stein2004efficient} of the deblurred event images.

The network predicts Euler angles, $(\psi, \beta, \phi)$, a translation, $T$, and the disparity of each pixel, $d_i$. The disparities are generated using the same encoder-decoder architecture as in the flow network, except that the final activation function is a sigmoid, scaled by the image width. The pose shares the encoder network with the disparity, and is generated by strided convolutions which reduce the spatial dimension from $16\times 16$ to $1\times 1$ with 6 channels.

\subsubsection{Temporal Reprojection Loss}
\label{sec:egomotion_loss}
Given the network output, the intrinsics of the camera, $K$, and the baseline between the two cameras, $b$, the optical flow, $(u_i,v_i)$ of each event at pixel location $(x_i,y_i)$ is:
\begin{align}
\begin{pmatrix}x^*_i\\y^*_i\end{pmatrix}=&K\pi\left(R\frac{fb}{d_i}K^{-1}\begin{pmatrix}x_i\\y_i\\1\end{pmatrix}+T\right)\\
\begin{pmatrix}u_i\\v_i\end{pmatrix}=&\frac{1}{B-1}\left(\begin{pmatrix}x^*_i\\y^*_i\end{pmatrix}-\begin{pmatrix}x_i\\y_i\end{pmatrix}\right)
\end{align}
where $f$ is the focal length of the camera, $R$ is the rotation matrix corresponding to $(\psi, \beta, \phi)$ and $\pi$ is the projection function: $\pi\left(\begin{pmatrix}X & Y & Z\end{pmatrix}^T\right)=\begin{pmatrix}\frac{X}{Z} & \frac{Y}{Z}\end{pmatrix}^T$. Note that, as the network only sees the discretized volume at the input, it does not know the size of the time window. As a result, the optical flow we compute is in terms of pixels/bin, where $B$ is the number of bins used to generate the input volume. The optical flow is then inserted into \eqref{eq:event_prop} for the loss.

\subsubsection{Stereo Disparity Loss}
\label{sec:disp_loss}
From the optical flow, we can deblur the events from the left and right camera using \eqref{eq:event_prop}, and generate a pair of event images, corresponding to the number of events at each pixel after deblurring. Given correct flow, these images represent the edge maps of the corresponding grayscale image, over which we can apply a photometric loss. However, the number of events between the two cameras may also differ, and so we apply a photometric loss on the census transforms~\cite{zabih1994non} of the images. For a given window width, $W$, we encode each pixel with a $W^2$ length vector, where each element is the sign of the difference between the pixel and each neighbor inside the window. For the left event volume, the right census transform is warped to the left camera using the left predicted disparities, and we apply a Charbonnier loss \eqref{eq:charbonnier} on the difference between the two images, and vice versa for the right.

In addition, we apply a left-right consistency loss between the two predicted disparities, as defined by \cite{godard2017unsupervised}.

Finally, we apply a local smoothness regularizer to the disparity, as in \eqref{eq:smoothness}.

The total loss for the SFM model is:
\begin{align}
\mathcal{L}_{SFM}=&\mathcal{L}_{temporal} + \lambda_2\mathcal{L}_{stereo} + \nonumber\\
&\lambda_3\mathcal{L}_{consistency} + \lambda_4\mathcal{L}_{smoothness}
\end{align}

%% file: tex/DeblurPic.tex
\begin{figure}[t]
    \centering
   	\includegraphics[width=0.15\textwidth]{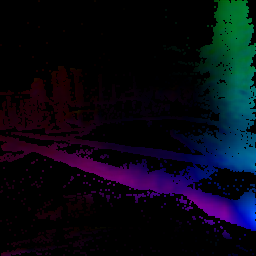}
   	\includegraphics[width=0.15\textwidth]{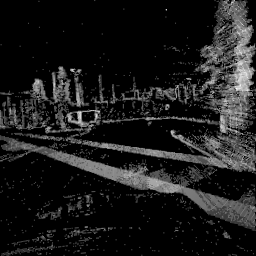}
   	\includegraphics[width=0.15\textwidth]{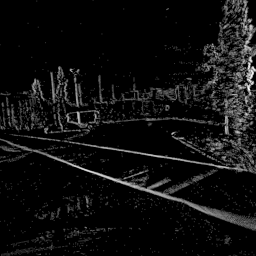}
    \caption{Our network learns to predict motion from motion blur by predicting optical flow or egomotion and depth (left) from a set of input, blurry, events (center), and minimizing the amount of motion blur after deblurring with the predicted motion to produce the deblurred image (right). Best viewed in color.}
    \label{fig:deblurring}
\end{figure}

%% file: tex/FlowResultsTable.tex
\begin{table*}[t!]
\centering
\begin{tabular}{ccccccccc} 
\hline
 & \multicolumn{2}{c}{outdoor day1} & \multicolumn{2}{c}{indoor flying1} & \multicolumn{2}{c}{indoor flying2} & \multicolumn{2}{c}{indoor flying3}\\
 

dt=1 frame& AEE & $\%$ Outlier & AEE & $\%$ Outlier & AEE & $\%$ Outlier & AEE & $\%$ Outlier\\

\hline\\[-1ex]
Ours & \textbf{0.32} & \textbf{0.0} & 0.58 & \textbf{0.0} & 1.02 & 4.0 & 0.87 & 3.0\\
EV-FlowNet
&  0.49 & 0.2 & 1.03  & 2.2 & 1.72 & 15.1  & 1.53 & 11.9\\
UnFlow
& 0.97  & 1.6  & 0.50  &  0.1 &  0.70 & 1.0 & 0.55 & \textbf{0.0}\\
ECN$_{\text{masked}}$
& 0.36 & 0.2 & \textbf{0.20$^*$} & \textbf{0.0$^*$} & \textbf{0.24$^*$} & \textbf{0.0$^*$} & \textbf{0.21$^*$} & \textbf{0.0$^*$}\\
\hline\\[-1ex]
& \multicolumn{2}{c}{outdoor day1} & \multicolumn{2}{c}{indoor flying1} & \multicolumn{2}{c}{indoor flying2} & \multicolumn{2}{c}{indoor flying3}\\
dt=4 frames & AEE & \% Outlier & AEE & \% Outlier & AEE & \% Outlier & AEE & \% Outlier\\
\hline\\[-1ex]
Ours & 1.30 & 9.7 & \textbf{2.18} & \textbf{24.2} & \textbf{3.85} & 46.8 & 3.18 & 47.8\\
EV-FlowNet
& \textbf{1.23}  &  \textbf{7.3} & 2.25  & 24.7 & 4.05  & \textbf{45.3} & 3.45 & 39.7\\
UnFlow
& 2.95  & 40.0 & 3.81 & 56.1 & 6.22 & 79.5 & \textbf{1.96} & \textbf{18.2} \\
ECN$_{\text{masked}}$
& - & - & - & - & - & - & - & -\\
\end{tabular}
\caption{Quantitative evaluation of our optical flow network compared to EV-FlowNet, UnFlow and ECN. For each sequence, Average Endpoint Error (AEE) is computed in pixels, \% Outlier is computed as the percent of points with AEE $<$ 3 pix. dt=1 is computed with a time window between two successive grayscale frames, dt=4 is between four grayscale frames. \\$^*$Evaluated on training data.}
\label{tab:flow_results}
\end{table*}

%% file: tex/ScaledDepthTable.tex
\begin{table}[t]
\centering
 \begin{tabular}{cccc} 
 \hline\\[-1ex]
  & \multicolumn{3}{c}{Average depth Error (m)} \\
  Threshold distance & 10m & 20m & 30m \\ [0.5ex] 
 \hline
 \multicolumn{4}{c}{outdoor$\_$day1}\\\hline
 Ours & \textbf{2.72} & \textbf{3.84} & \textbf{4.40} \\ 
 Monodepth & 3.44 & 7.02 & 10.03 \\\hline
 \multicolumn{4}{c}{outdoor$\_$night1}\\\hline
 Ours & \textbf{3.13} & \textbf{4.02} & \textbf{4.89}\\
 Monodepth & 3.49 & 6.33 & 9.31 \\\hline
 \multicolumn{4}{c}{outdoor$\_$night2}\\\hline
 Ours & \textbf{2.19} & \textbf{3.15} & \textbf{3.92} \\
 Monodepth & 5.15 & 7.8 & 10.03 \\\hline
 \multicolumn{4}{c}{outdoor$\_$night3}\\\hline
 Ours & \textbf{2.86} & \textbf{4.46} & \textbf{5.05} \\
 Monodepth &  4.67 & 8.96 & 13.36
\end{tabular}
\caption{Quantitative evaluation of our depth network compared to Monodepth~\cite{godard2017unsupervised}. The average depth error is provided for all points in the ground truth up to 10m, 20m and 30m, with at least one event.}
\label{tab:depth_results}
\end{table}

%% file: tex/Experiments.tex
\section{Experiments}
\label{sec:experiments}
\subsection{Implementation Details}
We train two networks on the full outdoor$\_$day2 sequence from MVSEC~\cite{zhu2018multi}, which consists of 11 mins of stereo event data driving through public roads. At training, each input consists of $N=30000$ events, which are converted into discretized event volumes with resolution 256x256 (centrally cropped) and $B=9$ bins. The weights for each loss are: $\{\lambda_1, \lambda_2, \lambda_3, \lambda_4\}=\{1.0, 1.0, 0.1, 0.2\}$.

\subsection{Optical Flow Evaluation}
We tested our optical flow network on the indoor$\_$flying and outdoor$\_$day sequences from MVSEC, with the ground truth provided by~\cite{zhu2018ev}. Flow predictions were generated at each grayscale frame timestamp, and scaled to be the displacement for the duration of 1 grayscale frame (dt=1) and 4 grayscale frames (dt=1), separately. For the outdoor$\_$day sequence, each set of input events was fixed at 30000, while for indoor$\_$flying, 15000 events were used due to the larger motion in the scene.

For comparison against the ground truth, we convert the output of the network, $(u,v)$, from units of pixels/bin into units of pixel displacement with the following: $(\hat{u}, \hat{v})=(u,v) \times (B-1) \times dt/ (t_N-t_0)$, where $dt$ is the test time window size.

We present the average endpoint error (AEE), and the percentage of points with AEE greater than 3 pixels, over pixels with valid ground truth flow and at least one event. These results can be found in Tab. \ref{tab:flow_results}, where we compare our results against EV-FlowNet~\cite{zhu2018ev}, the grayscale UnFlow~\cite{meister2017unflow}, and ECN~\cite{ye2018unsupervised}. However, we would like to note that most of the results provided by ECN~\cite{ye2018unsupervised} are evaluated on training data.
\subsection{Egomotion Evaluation}
We evaluate our ego-motion estimation network on the outdoor$\_$day1 sequence from MVSEC. We were only able to achieve reasonable results for outdoor$\_$day1, as the egomotion network did not generalize as well for the indoor$\_$flying sequences, and failed when observing fluorescent lights in the outdoor$\_$night sequences. This is discussed further in the results in Sec.~\ref{sec:results}.

As there is currently no public code to the extent of our knowledge for unsupervised deep SFM methods with a stereo loss, we compare our ego-motion results against SFMLearner by Zhou et al.~\cite{zhou2017unsupervised}, which learns egomotion and depth from monocular grayscale images, while acknowledging that our loss has access to an additional stereo image at training time. We train the SFMLearner models on the VI-Sensor images from the outdoor$\_$day2 sequence, once again cropping out the hood of the car. These images are of a higher resolution than the DAVIS images, but are from the same scene, and so should generalize as well as training on the DAVIS images. The model is trained from scratch for 100k iterations. As the translation predicted by SFMLearner is only up to a scale factor, we present errors in terms of angular error between both the predicted translation and rotations.

The relative pose errors (RPE) and relative rotation errors (RRE) are computed as: RPE $=\arccos\left(\frac{t_{\text{pred}}\cdot t_{\text{gt}}}{\|t_{\text{pred}}\|_2\|t_{\text{gt}}\|_2}\right)$, RRE $=\|\text{logm}(R_{\text{pred}}^TR_{\text{gt}})\|_2$, 
where $R_{\text{pred}}$ is the rotation matrix corresponding to the Euler angles output from each network, and logm is the matrix logarithm.
\input{tex/PoseResultsTable.tex}
\input{tex/TrajectoryFig.tex}
\subsection{Depth Network Evaluation}
We compare our depth results against Monodepth~\cite{godard2017unsupervised}, which learns monocular disparities from a stereo pair at training time, with an additional left-right consistency loss. As the DAVIS stereo grayscale images are not time synchronized, we once again train on the cropped VI-Sensor images. The model is trained for 50 epochs, and we provide depth errors for points with thresholds up to 10m, 20m and 30m in the ground truth and with at least one event. As the results from ECN are up to a scale and only provide relative depth results, we do not include them in our comparison.


%% file: tex/PoseResultsTable.tex
\begin{table}[t]
\centering
 \begin{tabular}{ccc} 
 \hline\\[-1ex]
  & ARPE (deg) & ARRE (rad)\\
 \hline \\[-1ex]
 Ours & \textbf{7.74} & \textbf{0.00867} \\ 
 SFM Learner~\cite{zhou2017unsupervised} & 16.27 &  0.00939 \\
\end{tabular}
\caption{Quantitative evaluation of our egomotion network compared to SFM Learner. ARPE: Average Relative Pose Error. ARRE: Average Relative Rotation Error.}
\label{tab:egomotion_results}
\end{table}

%% file: tex/TrajectoryFig.tex
\begin{figure}
    \centering
    \includegraphics[width=0.3\textwidth]{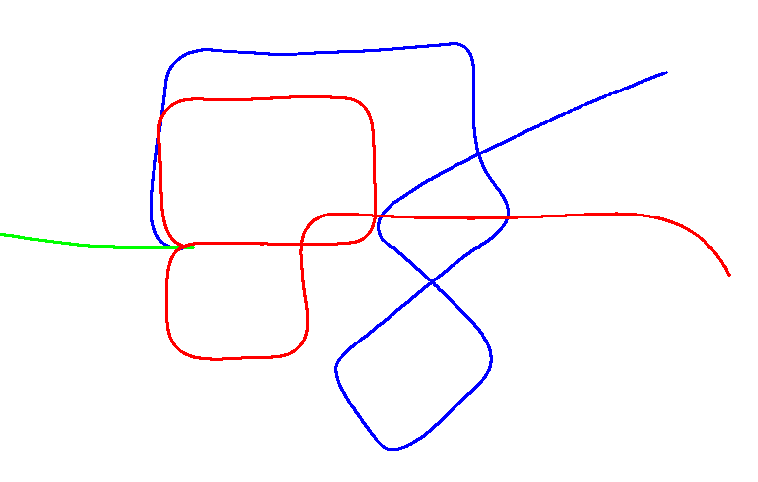}
    \caption{Estimated trajectory on outdoor$\_$day1 generated by concatenating egomotion predictions. Red: GT, Blue: Ours, Green: SFMLearner with GT scale.}
    \label{fig:trajectory}
\end{figure}

%% file: tex/Results.tex
\section{Results}
\label{sec:results}
\subsection{Optical Flow}
From the quantitative results in Tab.~\ref{tab:flow_results}, we can see that our method outperforms EV-FlowNet in almost all experiments, and nears the performance of UnFlow on the short 1 frame sequences. We also outperform ECN in the outdoor$\_$day1 sequence. We perform worse than ECN in the other sequences, but this is likely because these were in the training set for ECN. Qualitative results from these experiments can also be found in Fig.~\ref{fig:qualitative}.

In general, we have found that our network generalizes to a number of very different and challenging scenes, including those with very fast motions and dark environments. A few examples of this can be found in Fig.~\ref{fig:generalization}. We believe this is because the events do not have the fine grained intensity information at each pixel of traditional images, and so there is less redundant data for the network to overfit. However, our network does still struggle with events that are generated not as a result of motion, e.g. when there is a flashing light in the scene.

\subsection{Egomotion}
Our model trained on outdoor$\_$day2 was able to generalize well to outdoor$\_$day1, even though the environment is changed significantly from an outdoor residential environment to a closed office park area. In Tab.~\ref{tab:depth_results}, we show that our relative pose and rotation errors are significantly better than that of SFM-Learner, although this must be at least partially credited to the fact that our network has access to stereo images at training time. In addition, we show a section of the recovered trajectory in Fig.~\ref{fig:trajectory}. Due to the change in scene between outdoor$\_$day1 and outdoor$\_$day2, the network overestimates the scale of the trajectory, but is able to mostly accurately capture the rotation, and so the shape of the trajectory. SFM-Learner, on the other hand, consistently underestimates the rotation, and so diverges very quickly.

Unlike the flow network, both the egomotion and depth networks tended to memorize more of the scene, and as a result were unable to generalize well to sequences such as indoor$\_$flying. However, this still has valuable applications in operations where the environment does not vary significantly, such as geo-fenced autonomous driving applications. 

In addition, as the network was only trained on driving sequences, we were unable to achieve good egomotion generalization to the outdoor$\_$night sequences. We found that this was due to the fluorescent lamps found at night, which generated many spurious events due to their flashing that were not related to motion in the scene. As our egomotion network takes in global information in the scene, it tended to perceive these flashing lights as events generated by camera motion, and as a result generated an erroneous egomotion estimate. Future work to filter these kinds of anomalies out would be necessary for these networks to perform well. For example, if the rate of the flashing is known a-priori, the lights can be simply filtered by detecting events generated at the desired frequency.

\subsection{Depth}
Our depth model was able to produce good results for all of the driving sequences, although it is unable to generalize to the flying sequences. This is likely because the network must memorize some concept of metric scale, which cannot generalize to completely different scenes. We outperform Monodepth in all of the sequences, which is likely because the events do not have intensity information, so the network is forced to learn geometric properties of objects. In addition, the network generalizes well even in the face of significant noise at night, although flashing lights cause the network to predict very close depths, such as Fig.~\ref{fig:lights}.

\input{tex/LightEx.tex}

%% file: tex/LightEx.tex
\begin{figure}[t]
    \centering
   	\includegraphics[width=0.15\textwidth]{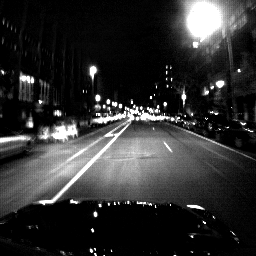}
   	\includegraphics[width=0.15\textwidth]{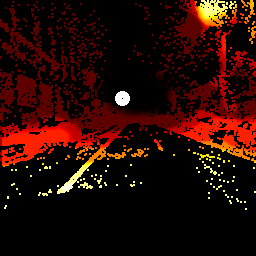}
    \caption{Failure case of our depth network, the flashing street light is detected as a very close object due to spurious events.}
    \label{fig:lights}
\end{figure}

%% file: tex/FlowFigs.tex
\begin{figure*}[t]
    \centering
   \includegraphics[width=0.14\textwidth]{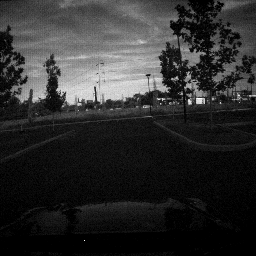}\;\;\;
     \includegraphics[width=0.14\textwidth]{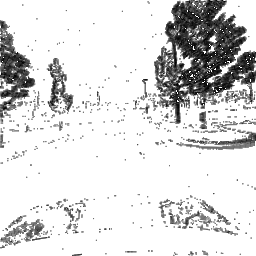}\;\;\;
     \includegraphics[width=0.14\textwidth]{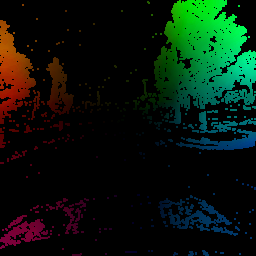}\;\;\;
     \includegraphics[width=0.14\textwidth]{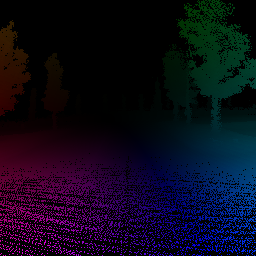}\;\;\;
     
     \includegraphics[width=0.14\textwidth]{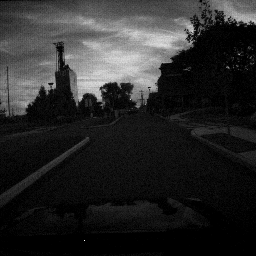}\;\;\;
     \includegraphics[width=0.14\textwidth]{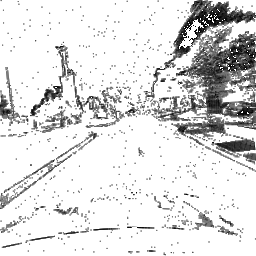}\;\;\;
     \includegraphics[width=0.14\textwidth]{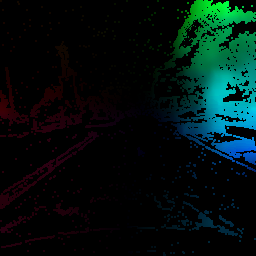}\;\;\;
     \includegraphics[width=0.14\textwidth]{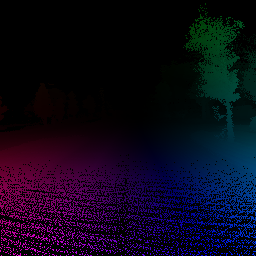}\;\;\;
     
      \includegraphics[width=0.14\textwidth]{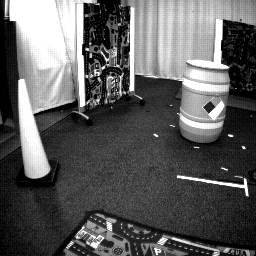}\;\;\;
     \includegraphics[width=0.14\textwidth]{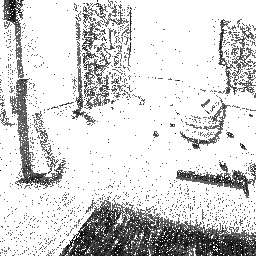}\;\;\;
     \includegraphics[width=0.14\textwidth]{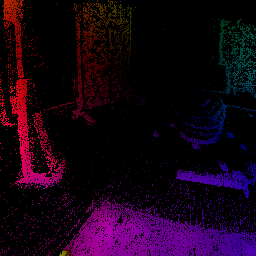}\;\;\;
     \includegraphics[width=0.14\textwidth]{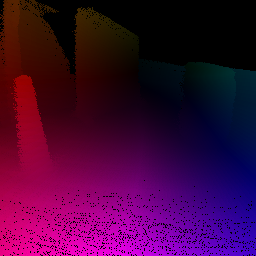}\;\;\;
     
     \includegraphics[width=0.14\textwidth]{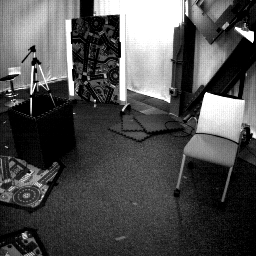}\;\;\;
     \includegraphics[width=0.14\textwidth]{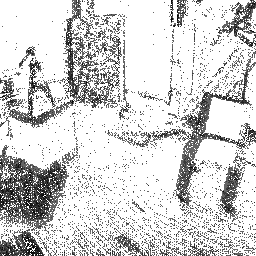}\;\;\;
     \includegraphics[width=0.14\textwidth]{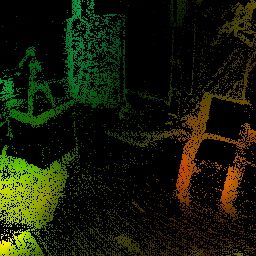}\;\;\;
     \includegraphics[width=0.14\textwidth]{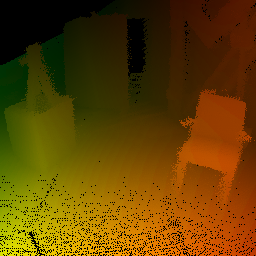}\;\;
\break \break     
	 \includegraphics[width=0.14\textwidth]{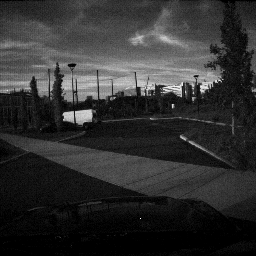}\;\;\;
     \includegraphics[width=0.14\textwidth]{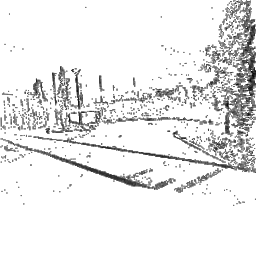}\;\;\;
     \includegraphics[width=0.14\textwidth]{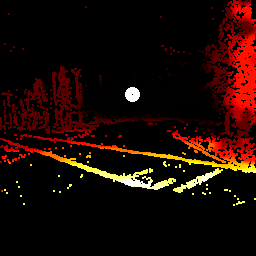}\;\;\;
     \includegraphics[width=0.14\textwidth]{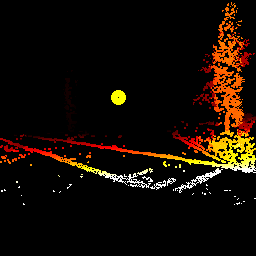}\;\;\;
     
   	 \includegraphics[width=0.14\textwidth]{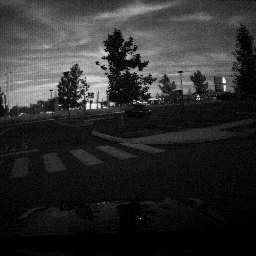}\;\;\;
     \includegraphics[width=0.14\textwidth]{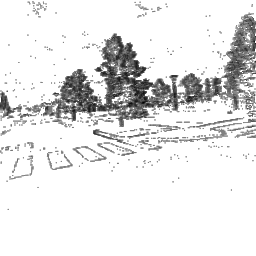}\;\;\;
     \includegraphics[width=0.14\textwidth]{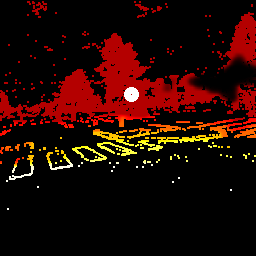}\;\;\;
     \includegraphics[width=0.14\textwidth]{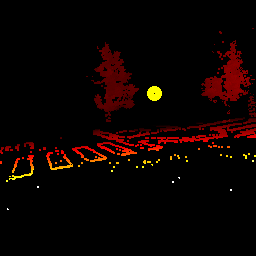}\;\;\;
     
	 \includegraphics[width=0.14\textwidth]{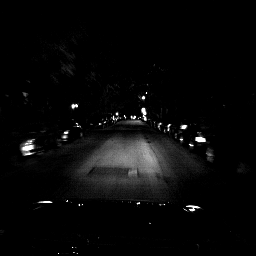}\;\;\;
     \includegraphics[width=0.14\textwidth]{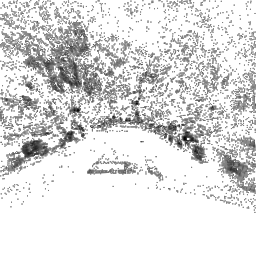}\;\;\;
     \includegraphics[width=0.14\textwidth]{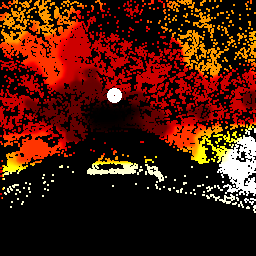}\;\;\;
     \includegraphics[width=0.14\textwidth]{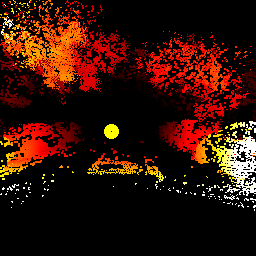}\;\;\;
     
  	 \includegraphics[width=0.14\textwidth]{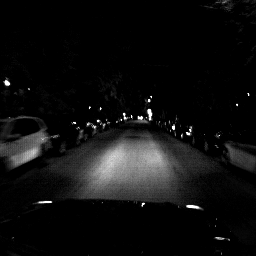}\;\;\;
     \includegraphics[width=0.14\textwidth]{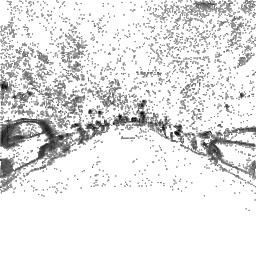}\;\;\;
     \includegraphics[width=0.14\textwidth]{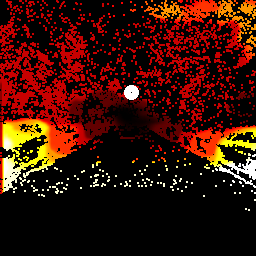}\;\;\;
     \includegraphics[width=0.14\textwidth]{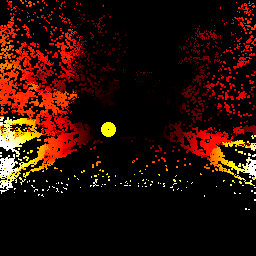}\;\;\;     
     \caption{Qualitative outputs from the optical flow and egomotion and depth network on the indoor$\_$flying, outdoor$\_$day and outdoor$\_$night sequences. From left to right: Grayscale image, event image, depth prediction with heading direction, ground truth with heading direction. Top four are flow results, bottom four are depth results. For depth, closer is brighter. Heading direction is drawn as a circle.}
     \label{fig:qualitative}
\end{figure*}